\def\year{2020}\relax
\documentclass[letterpaper]{article} 
\usepackage{aaai20}  
\usepackage{times}  
\usepackage{helvet} 
\usepackage{courier}  
\usepackage[hyphens]{url}  
\usepackage{graphicx} 
\usepackage{graphicx}  
\usepackage[dvipsnames]{xcolor}
\usepackage{amsfonts}
\usepackage{amsmath}
\usepackage{arydshln}
\usepackage{subcaption}
\usepackage{multirow}

\newcommand{\ie}{\textit{i}.\textit{e}.}
\newcommand{\eg}{\textit{e}.\textit{g}.}

\def\Baseline{\textit{Baseline}}
\def\OnPG{\textit{OnPG}}
\def\OffPG{\textit{OffPG}}
\newcommand{\BaselinePlus}[1]{\textit{Baseline$+(#1)$}}

\definecolor{myblue}{RGB}{10, 10, 100}
\definecolor{myred}{RGB}{100, 10, 10}

\title{Reinforcing an Image Caption Generator Using Off-Line Human Feedback}
\author{Paul Hongsuck Seo\textsuperscript{\rm 1,3}\thanks{Work done during internship at Google.} \hspace{0.2cm} Piyush Sharma\textsuperscript{\rm 2} \hspace{0.2cm}Tomer Levinboim\textsuperscript{\rm 2}  \hspace{0.2cm} Bohyung Han\textsuperscript{\rm 3} \hspace{0.2cm} Radu Soricut\textsuperscript{\rm 2}\\
\textsuperscript{\rm 1}Computer Vision Lab., POSTECH, Korea\\
\textsuperscript{\rm 2}Google Research, USA\\
\textsuperscript{\rm 3}Computer Vision Lab., ECE, \& ASRI, Seoul National University, Korea\\
hsseo@postech.ac.kr, \{piyushsharma, tomerl, rsoricut\}@google.com, bhhan@snu.ac.kr
}
\begin{document}

\maketitle

\begin{abstract}
Human ratings are currently the most accurate way to assess the quality of an image captioning model, yet most often the only used outcome of an expensive human rating evaluation is a few overall statistics over the evaluation dataset.
In this paper, we show that the signal from {\em instance-level} human caption ratings can be leveraged to improve captioning models, even when the amount of caption ratings is several orders of magnitude less than the caption training data.
We employ a policy gradient method to maximize the human ratings as rewards in an off-policy reinforcement learning setting, where policy gradients are estimated by samples from a distribution that focuses on the captions in a caption ratings dataset.
Our empirical evidence indicates that the proposed method learns to generalize the human raters' judgments to a previously unseen set of images, as judged by a different set of human judges, and additionally on a different, multi-dimensional side-by-side human evaluation procedure.
\end{abstract}

\section{Introduction}

Image captioning is the task of automatically generating fluent natural language descriptions for an input image.
However, measuring the quality of generated captions in an automatic manner is a challenging and yet-unsolved task;
therefore, human evaluations are often required to assess the complex semantic relationships between a visual scene and a generated caption~\cite{sharma2018conceptual,cui2018learning,zhao2019informative}.
As a result, there is a mismatch between the training objective of the captioning models and their final evaluation criteria.
The most simple and frequently-used training objective is maximum likelihood estimation (MLE) \cite{vinyals2015show,mun2017text,lu2017knowing,lu2018neural,changpinyo2019decoupled},
while other approaches make use of handcrafted evaluation metrics, such as
CIDEr~\cite{vedantam2015cider},
to optimize model parameters using reinforcement learning (RL) \cite{rennie2017self,ding2017coldstart,anderson2018bottom,qin2019look}.
However, these surrogate objectives capture only limited aspects of caption quality, and often fail to guide the training procedure towards models capable of producing outputs that are highly-rated by human evaluators.

\begin{figure}[t]
\centering
    \begin{subfigure}[m]{0.45\linewidth}
        \centering
        \includegraphics[width=1\linewidth]{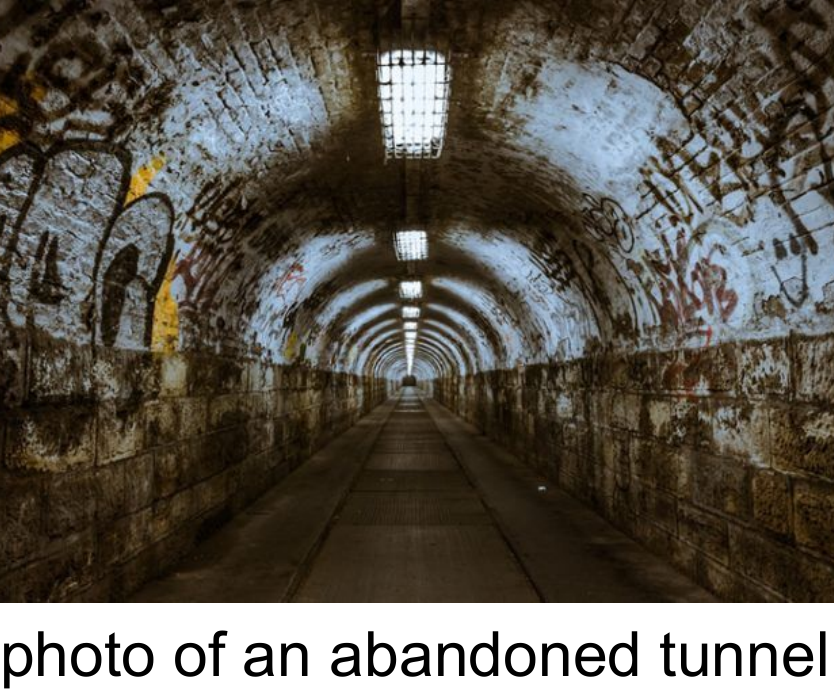}
        \caption{Image caption}
    \end{subfigure}
    \hspace{0.5cm}
    \begin{subfigure}[m]{0.45\linewidth}
        \centering
        \includegraphics[width=1\linewidth]{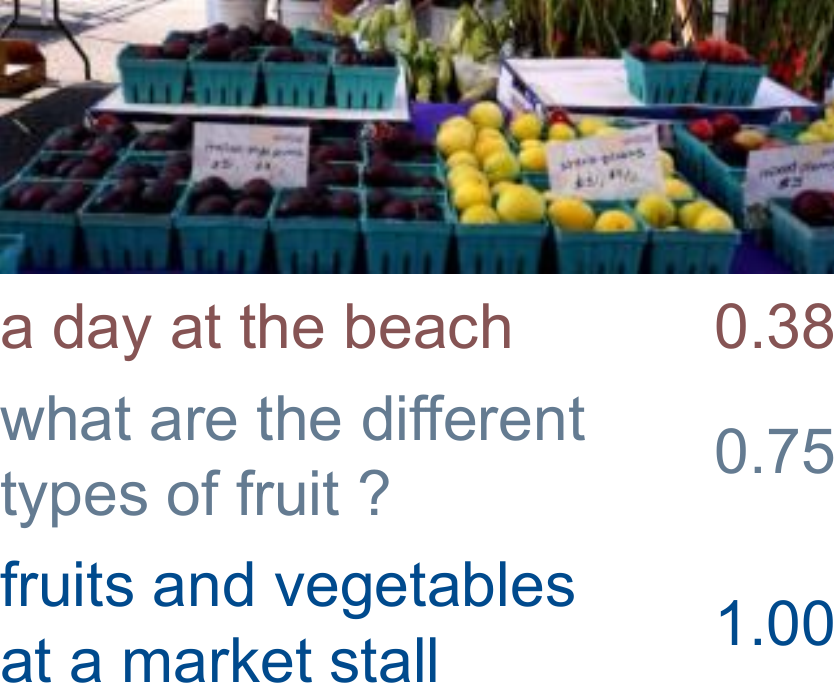}
        \caption{Captions ratings}
        \label{fig:OIC_CR_dataset}
    \end{subfigure}%
    \caption{
    Examples from an image caption dataset~\cite{sharma2018conceptual} and a caption-ratings dataset~\cite{levinboim2019quality}.
    (a) Images in the caption dataset are annotated with ground-truth captions written by humans.
    (b) Captions in the caption ratings dataset are generated by trained models and scored in [0, 1] (0: worst, 1: best) by human raters.
    }
    \label{fig:datasets}
\end{figure}

As a result of the need to understand the performance of the current models, human evaluation studies for measuring caption quality are frequently reported in the literature ~\cite{sharma2018conceptual,forbes2019neural,dognin2019adversarial,zhao2019informative}.
In addition to an aggregate model performance, such human evaluation studies also produce a valuable by-product: a dataset of model-generated image captions with human annotated quality labels, as shown in Figure~\ref{fig:OIC_CR_dataset}.
We argue that such a by-product, henceforth called a caption ratings dataset, can be successfully used to improve the quality of image captioning models, for several reasons.
First, optimizing based on instance-level human judgments of caption quality represent a closer-to-truth objective for image captioning: generating more captions judged as good but fewer ones rated as poor by human raters.
Second, while having highly-rated captions as positive examples 
(\ie, how good captions may look like), a caption ratings dataset also contains captions that are highly-scored by a model but annotated as negative examples (\ie, how model-favored yet bad captions look like), which intuitively should be a useful signal for correcting common model biases.
To the best of our knowledge, our work is the first to propose using human caption ratings directly for training captioning models.
%

Our goal is to leverage the signals from a pre-collected caption ratings dataset \cite{levinboim2019quality} for training an image captioning model.
We propose a method based on policy gradient, where the human ratings are considered as rewards for generating captions (seen as taking actions) in an RL framework.
Since the dataset provides ratings only for a small set of images and captions, we do not have a generic reward function for random image-caption pairs.
Therefore, it is not straightforward to apply policy gradient method that requires a reward for randomly sampled captions.
To address this challenge, we use an off-policy technique and force the network to sample captions for which ratings are available in the dataset.
We evaluate the effectiveness of our method using human evaluation studies on the T2 test set used for the Conceptual Captions Challenge\footnote{http://www.conceptualcaptions.com/challenge}, using both a similar human evaluation methodology and an additional, multi-dimensional side-by-side human evaluation strategy.
Additionally, the human raters in our evaluation study are different from the ones that provided the caption ratings in \cite{levinboim2019quality}, thereby ensuring that the results are independent of using a specific human-evaluator pool.
The results of our human evaluations indicate that the proposed method improves the image captioning quality, by effectively leveraging \emph{both} the positive and negative signals from the captions ratings dataset.


The main contributions of this paper are the following:
\begin{itemize}
\item We propose to train captioning models using human ratings produced during evaluations of previous models. 
\item We propose an off-policy policy gradient method to cope with the sparsity in available caption ratings. 
\item We present a set of experiments using human evaluations that demonstrates the effectiveness of our approach.
\end{itemize}


\section{Related Work}
\label{sec:related_work}
There have been multiple attempts to define metrics that evaluate the quality of generated captions.
Several studies proposed automatic metrics using ground-truth captions. 
A few of them are adopted from machine translation community and are based on $n$-gram matches between ground-truth and generated captions; BLEU~\cite{papineni2002bleu} and ROUGE~\cite{lin2004rouge} measures precision and recall based on $n$-gram matches, respectively, while METEOR~\cite{banerjee2005meteor} incorporates alignments between $n$-gram matches.
In the context of evaluating image caption quality specifically, CIDEr~\cite{vedantam2015cider} and SPICE~\cite{anderson2016spice} utilize more corpus-level and semantic signals to measure matches between generated and ground-truth captions. 
Aside from these handcrafted metrics, a recent study proposes to learn an automatic metric from a captioning dataset~\cite{cui2018learning}, while another uses semantic similarity between object labels identified
in the image and the words in the caption~\cite{madhyastha2019vifidel}. 

To overcome the limitations imposed by the automatic metrics, several studies evaluate their models using human judgments~\cite{sharma2018conceptual,zhao2019informative,dognin2019adversarial,forbes2019neural}.
However, none of them utilizes the human-rated captions in the model evaluations.
In this work, we show how one can utilize such human-rated captions for training better captioning models.

MLE with ground-truth captions has been widely adopted as the standard surrogate objective for training~\cite{vinyals2015show,mun2017text,lu2017knowing,lu2018neural,changpinyo2019decoupled}.
Aside from this main thrust, an additional line of research is concerned with optimizing models that maximize some automatic evaluation metric(s) using RL, in an attempt to bridge the mismatch between the training objective and the evaluation criteria~\cite{rennie2017self,ding2017coldstart,anderson2018bottom,qin2019look}.
To our knowledge, this is the first study that proposes to optimize test-time scores of human judgment using a dataset generated by a human evaluation process.

Another line of related research is focused on learning from human feedback, which has been actively explored in the field of RL.
Some approaches use binary human feedback to update an agent~\cite{knox2012humans,amershi2014power,macglashan2017interactive} whereas approaches with preference-based RL take human feedback as preferences between two action/state trajectories~\cite{akrour2012april,wirth2016model,wirth2017survey}.
A common technique adopted in these methods is to learn an estimation model from human feedback to approximate the absent reward function~\cite{knox2012humans,christiano2017deep,ibarz2018reward}.
However, these approaches assume that the models receive human feedback iteratively in a training loop; in contrast, our approach uses the caption ratings in an off-line manner, simply as a pre-existing annotation dataset.
As a result, our method focuses on existing examples within the dataset, using an off-policy technique.



\begin{figure*}
    \centering
    \begin{tabular}{ccc}
        $\bigtriangledown_\theta \ln p_\theta(c^\mathrm{GT}|I)$ &
        $\tilde{r}(c_s|I)\bigtriangledown_\theta \ln p_\theta(c_s|I)$ &
        $\eta r(c'_s|I)\bigtriangledown_\theta \ln p_\theta(c'_s|I)$ \vspace{0.2cm}\\
        \hspace{0.4cm}
        \includegraphics[width=0.22\linewidth]{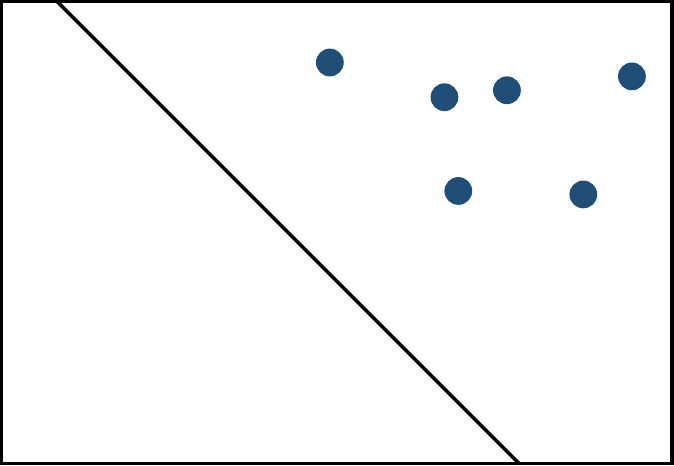} 
        \hspace{0.4cm} & 
        \hspace{0.4cm}
        \includegraphics[width=0.22\linewidth]{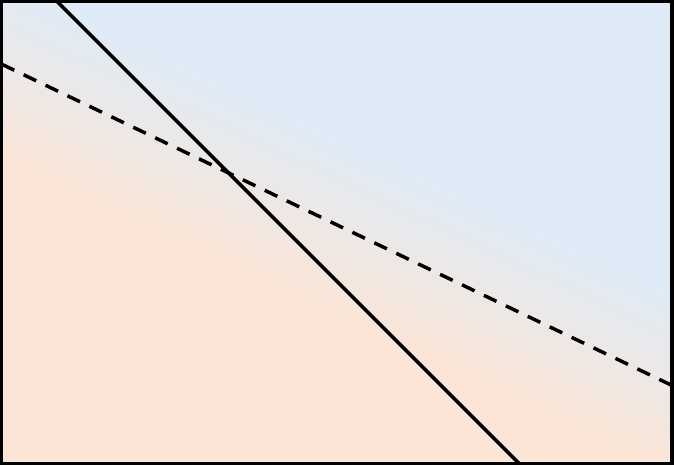} 
        \hspace{0.4cm} & 
        \hspace{0.4cm}
        \includegraphics[width=0.22\linewidth]{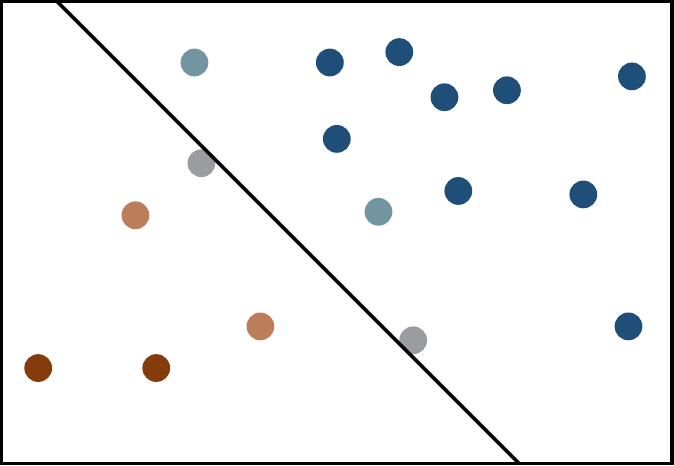}
        \hspace{0.4cm} \vspace{0.1cm}\\
        \multirow{2}{*}{(a) Maximum likelihood estimation} & (b) On-policy policy gradient & (c) Off-policy policy gradient \\
          & with rating estimator & with true human ratings \\
    \end{tabular}
    \caption{
    Illustration of training settings for three different methods.
    The 2D boxes represent the space of possible captions, and each dot is a caption with its color corresponding to human ratings. (Blue is high and red is low.)
    The solid line in each plot indicates the virtual boundary separating low-quality and high-quality captions in terms of human ratings.
    The color gradation in (b) represents learned rating estimates for captions, while the dashed line is the model's approximated virtual boundary between low- and high-quality estimates. In (c), $\eta=\frac{p_\theta(c'_s|I)}{q(c'_s|I)}$ denotes the importance weight for a sample $c'_s$.
    }
    \label{fig:methods_relation}
\end{figure*}

\section{Methods}
\label{sec:methods}

\subsection{Caption Ratings Dataset}
A sample in a caption ratings dataset is comprised of an image $I$, a machine-generated caption $c$, and a human judgment for the caption quality $r(c|I) \in \mathbb{R}$. 
For each image, multiple captions from several candidate models are available, some of which might be rated higher than others.
In the setup used in this paper, the low-rated captions serve as negative examples, because human annotators judged them as bad captions (see examples in Figure~\ref{fig:OIC_CR_dataset}).
$r(c|I)$ is possibly an aggregate of multiple ratings from different raters.
Section~\ref{sec:qe-dataset} provides more details of the caption ratings dataset that we employ.

We make a few observations that apply not only to image captioning, but more generally to the principle of generating annotations.
Although a human-ratings dataset is usually just a by-product of human evaluations for past models, such a dataset can be valuable for improving models (as we show in this paper).
There are several advantageous properties of a ratings dataset over traditional supervised-learning datasets.
First, obtaining ratings for automatically generated outputs is significantly cheaper than collecting ground-truth labels, because it requires less rater training and less time spent annotating.
Moreover, if human evaluation is performed anyway during a model's development cycle, there is no additional cost associated to using these annotations for further improving the model.
In addition to that, it is easy to capture consensus between multiple raters to reduce noise, \eg, by averaging their scores; it is completely non-trivial to achieve a similar effect from multiple ground-truth labels.
Last but not least, the examples with a negative rating score provide valuable training signals, as they explicitly penalize the mistakes that appear in model outputs with high-probability; this type of signal is completely lacking in traditional supervised-learning datasets.

\subsection{Reinforcing Caption Generator using Ratings}
Given a caption ratings dataset $\mathcal{D}$ with triplets $(I, c, r(c|I))$, our objective is to maximize the expected ratings of the output captions $\mathcal{J}(\theta)$, which is given by
\begin{align}
    \mathcal{J}(\theta)
    &=  \mathbb{E}_{I\sim p_\mathcal{D}(I),c\sim p_\theta(c|I)}[r(c|I)] \label{eq:objective} \nonumber \\
    &= \sum_{I} p_\mathcal{D}(I) \sum_{c} p_\theta(c|I) r(c|I),
\end{align}
where $p_\mathcal{D}(I)$ is the dataset distribution for $I$ and $p_\theta(c|I)$ is the conditional caption distribution estimated by a model parameterized by $\theta$.

Our objective in Eq.~\eqref{eq:objective} exactly aligns with the reward maximization of RL, and therefore we apply the techniques of RL by configuring the captioning model as the agent, the rating scores as the reward, the input images as the states, and the captions as the actions.
Specifically, we use a policy gradient method where an approximated policy gradient is computed using Monte-Carlo sampling,
\begin{align}
    \bigtriangledown_\theta \mathcal{J}_\mathrm{PG}(\theta) 
    &= \mathbb{E}_{\pi}[(r(c|I)-b)\bigtriangledown_\theta \ln p_\theta(c|I)] \nonumber\\
    &\approx \frac{1}{S} \sum_{s=1}^S (r(c_{s}|I_{s})-b)\bigtriangledown_\theta \ln p_\theta(c_{s}|I_{s}), \label{eq:pg}
\end{align}
where $\mathbb{E}_\pi$ represents $\mathbb{E}_{I\sim p_\mathcal{D}(I),c\sim p_\theta(c|I)}$, $I_s$ and $c_s$ are image and caption sampled from $p_\mathcal{D}(I)$ and $p_\theta(c|I)$, respectively, and $S$ is the number of samples.
In the above equations, we subtract a baseline $b$ from the rating score $r(c_{s}|I_{s})$ to reduce the variance of the estimator while keeping its original bias.

Although this formulation is straightforward, there remains a critical challenge to apply this technique to our task, since the dataset $\mathcal{D}$ contains only sparse information about $r(c|I)$ and true ratings for most captions are unknown.
Eq.~\eqref{eq:pg} requires the rating $r(c_s|I_s)$ for a randomly sampled caption which may not be present in the dataset $\mathcal{D}$.
In the rest of this section, we present two alternative techniques for this challenge, and discuss the advantages of one alternative versus the other.


\subsubsection{On-policy policy gradient with rating estimates}
One approach to address the sparsity of the rating function is to construct a caption quality estimator, while keeping the sampling process on-policy; this is the method adopted in, \eg,~\cite{knox2012humans,christiano2017deep,ibarz2018reward}.
Incidentally, it is also the expressed goal for the effort behind the caption ratings dataset in~\cite{levinboim2019quality} that we use in this work.

For this purpose, we train a rating estimator $\tilde{r}(c|I;\phi)$ parameterized by $\phi$, by minimizing mean squared error of the true rating scores for the image-caption pairs on the caption ratings dataset.
The trained estimator then replaces the true rating function $r(c_s|I_s)$ in Eq.~\eqref{eq:pg} and the estimated policy gradient is now:
\begin{equation}
    \bigtriangledown_\theta \mathcal{J}_\mathrm{PG}(\theta) \approx \frac{1}{S} \sum_{s=1}^S (\tilde{r}(c_{s}|I_{s};\phi)-b)\bigtriangledown_\theta \ln p_\theta(c_{s}|I_{s}). \label{eq:onpg}
\end{equation}

This technique allows to obtain rating estimates for any image-caption pairs, including ones that are not present in the dataset $\mathcal{D}$.
The training objective with Eq.~\eqref{eq:onpg} is now maximizing the expected rating estimate of captions.
This approach is effective only if the trained rating estimator generalizes well to unseen images and captions, and it is expected to be effective only to the extent to which the rating estimator performs well over the sampled search space.
In our work, we have observed artifacts of the ratings estimator that negatively impact the performance of this method, \eg, severely ill-formed captions for which the caption estimator had no training signal but
assigned high ratings.
We report results for this method in Section~\ref{sec:experiments}.

\subsubsection{Off-policy policy gradient with true ratings}
This second method takes an orthogonal approach to address the sparsity of the rating function.
We modify the sampling process in such a manner that it allows us to directly utilize the true ratings of the dataset (no estimation involved), while ensuring that the training procedure is not influenced by the captions whose true ratings are not available.
More precisely, we adopt an off-policy policy gradient technique that uses an alternative distribution $q(c|I)$, instead of the true policy distribution $p_\theta(c|I)$ for sampling.
The policy gradient in Eq.~\eqref{eq:pg} is approximated as follows:
\begin{align}
    \bigtriangledown_\theta \mathcal{J}_\mathrm{PG}(\theta) 
    =& \mathbb{E}_{\pi}[(r(c|I)-b)\bigtriangledown_\theta \ln p_\theta(c|I)] \\
    =& \mathbb{E}_{\beta}\left[\frac{p_\theta(c|I)}{q(c|I)} (r(c|I)-b)\bigtriangledown_\theta \ln p_\theta(c|I)\right] \nonumber \\
    \approx& \frac{1}{S} \sum_{s=1}^S \frac{p_\theta(c|I)}{q(c|I)} (r(c_{s}|I_{s})-b)\bigtriangledown_\theta \ln p_\theta(c_{s}|I_{s}),  \nonumber  \label{eq:offpg}
\end{align}
where $\mathbb{E}_\beta$ represents $\mathbb{E}_{I\sim p_\mathcal{D}(I),c\sim q(c|I)}$ with an alternative caption distribution $q(c|I)$, and $\frac{p_\theta(c|I)}{q(c|I)}$ represents the importance weight for sample caption $c_s$ and image $I_s$.
The alternative caption sampling distribution is defined as: 
\begin{equation}
    q(c|I) = (1-\epsilon)p_\mathcal{D}(c|I) + \epsilon U(c),
\end{equation}
where $p_\mathcal{D}(c|I)$ is the conditional caption distribution in the dataset $\mathcal{D}$, $U(\cdot)$ is the uniform distribution, and $\epsilon \ll 1$ is a small positive weight assigned to the uniform distribution.
In all experiments, we sample a single caption per image in the batch.
While captions that are not present in the dataset may still be sampled from $U(c)$, we assign a reward $b$ to these captions, in order to prevent incorrect contributions to the gradient computation.
In the policy gradient formulation, examples with reward value $b$ are considered to have no information, and their weight $r(c|I)-b=0$ cancels out the entire term corresponding to these examples.
Note that the off-policy methods enable experience replay, which is repeating previous experiences with known rewards.
In this view, this method is viewed as training a captioning model by replaying the experiences in the ratings dataset. 

\subsubsection{Curriculum learning}
As our training conditions, we assume the access to both a captioning dataset and a caption ratings dataset.
Under a curriculum learning procedure, we first train a model by MLE on the captioning dataset, and then fine-tune the model with the above methods using the caption ratings dataset.
To avoid overfitting during fine-tuning, we add the MLE loss on the captioning dataset as a regularization term.
Given the caption labeled dataset $\mathcal{D}_\mathrm{IC}$ and the caption ratings dataset $\mathcal{D}_\mathrm{CR}$, the final gradients w.r.t. the parameters are therefore computed as follows:
\begin{align}
    \bigtriangledown_\theta \mathcal{J}(\theta) &= \alpha \bigtriangledown_\theta \mathcal{J}_\mathrm{PG}(\theta; \mathcal{D}_\mathrm{CR}) + \bigtriangledown_\theta \mathcal{J}_\mathrm{MLE}(\theta; \mathcal{D}_\mathrm{IC}),
\end{align}
where $\mathcal{J}_\mathrm{MLE}$ is the average log-likelihood of ground-truth captions in $\mathcal{D}_\mathrm{IC}$, and $\alpha$ is a hyper-parameter that balances the regularization effect.

\subsection{Comparing two policy gradient methods}
Intuitively, the two policy gradient methods described in this section have strong relationships to MLE, since training signals are based on the gradients of caption log-likelihoods.
We illustrate the training settings of MLE and the two proposed methods in Figure~\ref{fig:methods_relation}.
%
In MLE, we train the model using positive captions only and treat all positive captions equally, as illustrated in Figure~\ref{fig:methods_relation}a: the parameters are updated by the gradients of log-likelihoods of ground-truth captions $c_\mathrm{GT}$.
The on-policy policy gradient method (Eq.~\eqref{eq:onpg}) instead computes the gradients of reward-weighted log-likelihoods of sample captions $c_s$ over all possible captions.
By sampling from the policy distribution (on-policy), we may sample captions whose true rating scores are not known (not in the dataset).
The on-policy method thus approximates the rating function by a rating estimator $\tilde{r}(c|I)$, depicted by the background gradient in Figure~\ref{fig:methods_relation}b.
However, the mismatch between the true rating function and the estimator (depicted by the gap between solid and dashed lines) can degenerate the quality of the resulting captioning model.
On the other hand, the off-policy method focuses on the captions with true rating scores in the dataset, by changing the sampling distribution.
In contrast to MLE, where each sample is viewed as equally correct and important, the off-policy method weights each caption by its rating, and therefore includes captions with negative feedback, as illustrated in Figure~\ref{fig:methods_relation}c.
Note that, in the off-policy method, the baseline determines the threshold for positive/negative feedback;
captions with ratings below the baseline are explicitly penalized, while the others are positively rewarded.



\section{Experiments}
\label{sec:experiments}
\subsection{Datasets}
\label{sec:datasets}
\subsubsection{Image captioning dataset}
In the experiments, we use Conceptual Captions~\cite{sharma2018conceptual}, a large-scale captioning dataset that consists of images crawled from the Internet, with captions derived from corresponding Alt-text labels on the webpages.
The training and validation splits have approximately 3.3M and 16K samples, respectively.

\subsubsection{Caption ratings dataset}
\label{sec:qe-dataset}
In our experiments, we use the Caption-Quality dataset~\cite{levinboim2019quality}, recently introduced for the purpose of training quality-estimation models for image captions.
We re-purpose this data as our caption ratings dataset $\mathcal{D}_\mathrm{CR}$.
The dataset is divided into training, validation and test splits containing approximately 130K, 7K and 7K rated captions, respectively.
Each image has an average of 4.5 captions (generated by different models that underwent evaluation evaluation).
The captions are individually rated by asking raters the question \textit{``Is this a good caption for the image?''}, with the answers \textit{``NO''} or \textit{``YES''} mapped to a 0 or 1 score, respectively.
Each image/caption pair is evaluated by 10 different human raters, and an average rating score per-caption is obtained by quantizing the resulting averages into a total of nine bins $\{0, \frac{1}{8} \dots \frac{7}{8}, 1\}$.

\subsubsection{Conceptual Captions Challenge T2 dataset}
To evaluate our models, we run human evaluation studies on the T2 test dataset used in the CVPR 2019 Conceptual Captions Challenge\footnote{http://www.conceptualcaptions.com/winners-and-data}.
The dataset contains 1K images sampled from the Open Images Dataset~\cite{kuznetsova2018open}.
Note that the images in the Caption-Quality dataset are also sampled from the Open Images Dataset, but using a disjoint split.
So there is no overlap between the caption ratings dataset $\mathcal{D}_\mathrm{CR}$ we use for training, and the T2 test set we use for evaluations.

\subsection{Experimental Settings}

\begin{figure}[t]
    \centering
    \begin{subfigure}[m]{\linewidth}
        \centering
        \includegraphics[width=0.9\linewidth]{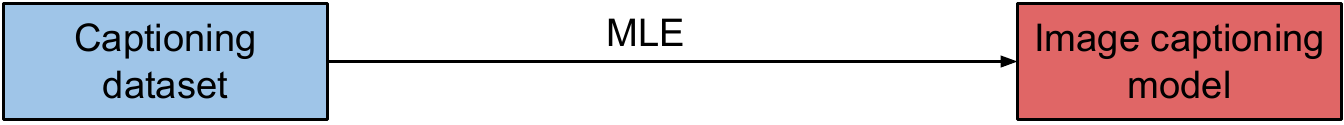}
        \label{fig:ref_training}
        \caption{Baseline}
    \end{subfigure} \\
    \vspace{0.2cm}
    \begin{subfigure}[m]{\linewidth}
        \centering
        \includegraphics[width=0.9\linewidth]{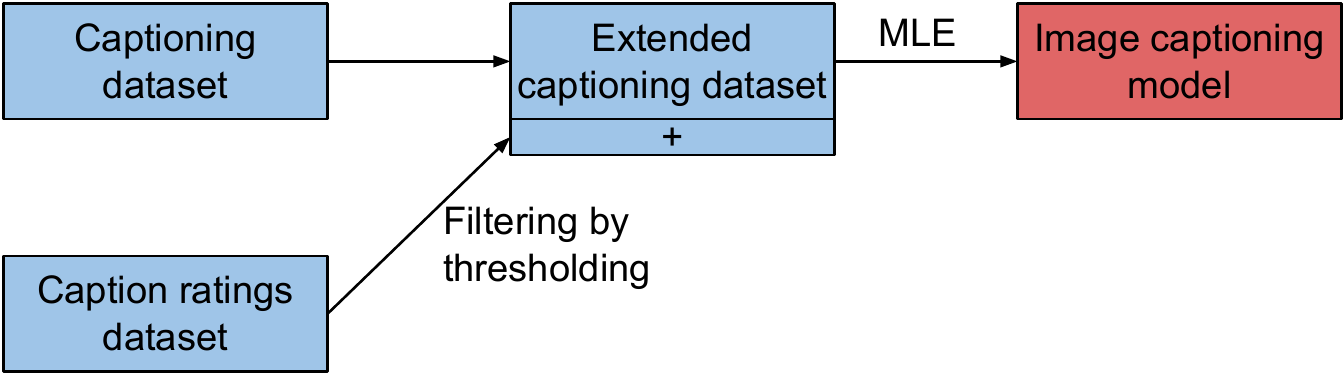}
        \caption{Baseline+}
        \label{fig:base_training}
    \end{subfigure} \\
    \vspace{0.2cm}
    \begin{subfigure}[m]{\linewidth}
        \centering
        \includegraphics[width=0.9\linewidth]{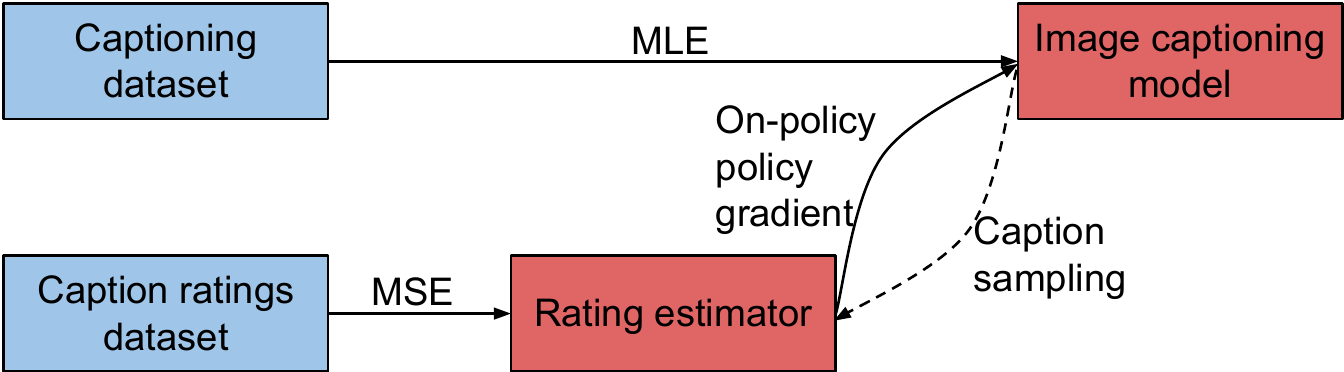}
        \caption{On-policy policy gradient with rating estimator (OnPG)}
        \label{fig:onpg_training}
    \end{subfigure} \\
    \vspace{0.2cm}
    \begin{subfigure}[m]{\linewidth}
        \centering
        \includegraphics[width=0.9\linewidth]{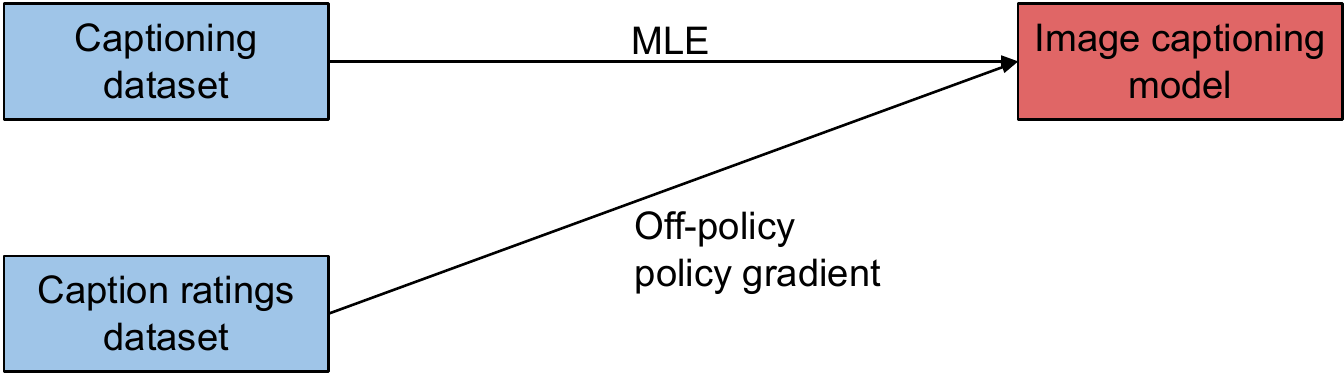}
        \caption{Off-policy policy gradient with true ratings (OffPG)}
        \label{fig:offpg_training}
    \end{subfigure}
    \caption{
    The training procedures for the different methods described. Blue and red boxes represent datasets and models, respectively. 
    MLE is maximum likelihood estimation and MSE means mean-squared error minimization.
    }
    \label{fig:training_fig}
\end{figure}

\renewcommand{\arraystretch}{1.1} 
\begin{table}[t]
    \centering
    \caption{Questions asked to raters in the two human evaluations. Type 'S' evaluation means single-caption rating. Type `SxS' evaluation means side-by-side caption rating.}
    \label{tab:questions}
    \scalebox{0.8}{
    \begin{tabular}{lcp{6cm}}
        Metric & Type & Question  \\ \hline\hline
        Goodness & S & Is this a good caption for the image? \\\hdashline
        Informativeness & SxS & Which caption provides more useful info \\
                        &     & for a person who cannot see this image? \\
        Correctness & SxS & Which caption has fewer mistakes? \\
        Fluency & SxS & Which caption has better language quality? \\
    \end{tabular}
    }
\end{table}

\subsubsection{Model architecture}
As the backbone model for image captioning we adopt the architecture described in~\cite{changpinyo2019decoupled}, since it provides the highest single-model score in the Conceptual Captions Challenge\footnote{As of Sept. 5, 2019.}.
Given an image, we extract two types of visual features: 1) ultra fine-grained semantic features using pretrained network~\cite{juan2019graph} from the entire image and 16 bounding boxes proposed by faster-RCNN~\cite{ren2015faster}, and 2) label embeddings of objects predicted by Google Cloud Vision API\footnote{http://cloud.google.com/vision}.
We use these features with an encoder-decoder Transformer Network~\cite{vaswani2017attention} to generate the captions.

In addition, we train a caption rating estimator for the \OnPG~method using the Caption-Quality dataset.
The rating estimator extracts the same types of visual features as the captioning model above, and embeds the input caption with a pretrained BERT encoder~\cite{devlin2018bert}.
We concatenate all these features after projecting into a common embedding space and predict the human ratings of the input image/caption pair.
To feed the generated captions from the captioning model directly into the rating estimator, we share the vocabulary (but not the token embeddings) between the two models.
We fix the pretrained image feature extraction modules in both models during training, as well as the BERT encoder of the rating estimator.
The rating estimator achieves a test performance that is close to the one reported (0.519 Spearman correlation) in \cite{levinboim2019quality}; however, as we will discuss further, its performance on the Caption-Quality test set does not transfer well to the needs of the \OnPG~method, which needs correct rating estimates for ill-formed captions as well.

\subsubsection{Baselines and proposed models}
We first train an MLE model as our baseline, trained on the Conceptual Captions training split alone.
We referred to this model as \Baseline.
For a baseline approach that utilizes (some of) the Caption-Quality data, we merge positively-rated captions from the Caption-Quality training split with the Conceptual Captions examples and finetune the baseline model.
We call this model \BaselinePlus{t}, where $t \in [0,1]$ is the rating threshold for the included positive captions.
We train models for two variants, $t\in\{0.5, 0.7\}$, which results in $\sim$72K and $\sim$51K additional (pseudo-)ground-truth captions, respectively.
Note that the \BaselinePlus{t} approaches attempt to make use of the same additional dataset as our two reinforced models, \OnPG~and \OffPG, but they need to exclude below-threshold captions due to the constraints in MLE.

In addition to the baselines, we train two reinforced models: one based on the on-policy policy gradient method with a rating estimator (\OnPG), and the other based on the off-policy policy gradient method with the true ratings (\OffPG).
The differences between the methods are shown in Figure~\ref{fig:training_fig}.

\subsubsection{Training details}
We train \Baseline{} using the Adam optimizer~\cite{kingma2014adam} on the training split of the Conceptual dataset for 3M iterations with the batch size of 4,096 and the learning rate of $3.2\times10^{-5}$.
The learning rate is warmed up for 20 epochs and exponentially decayed by a factor of 0.95 every 25 epochs.
\BaselinePlus{t} are obtained by fine-tuning \Baseline{} on the merged dataset for 1M iterations, with the learning rate of $3.2\times10^{-7}$ and the same decaying factor.
For \OnPG{}, because its memory footprint is increased significantly due to the additional parameters for the rating estimator, we reduce the batch size for training this model by a 0.25 factor; the value of $b$ in Eq.~\eqref{eq:pg} is set to the moving average of the rating estimates.
During \OffPG{} training, for each batch, we sample half of the examples from the Conceptual dataset and the other half from Caption-Quality dataset; $b$ is set to the average of the ratings in the dataset.

\begin{table}[t]
    \caption{Human evaluation single-caption results: Goodness scores for models (higher is better). Column $\bigtriangleup$ shows relative improvements over \Baseline.
    Note that all score increases of \Baseline+(t)~and \OnPG~are within the error range.}
    \label{tab:goodness}
    \centering
    \scalebox{0.798}{
    \begin{tabular}{l|cc|cc}
                            & \multicolumn{2}{c|}{Average}      & \multicolumn{2}{c}{Voting}    \\ 
                        & Goodness & $\bigtriangleup$  & Goodness & $\bigtriangleup$  \\ \hline\hline
        Baseline        & 66.23$\pm$0.60\%    & ---  & 66.30$\pm$1.49\% & ---                     \\\hdashline
        Baseline+ (0.5) & 66.68$\pm$0.61\%  & 0.45\%    & 66.50$\pm$1.48\% & 0.20\%                 \\
        Baseline+ (0.7) & 65.83$\pm$0.62\%	& -0.40\%	& 65.20$\pm$1.51\% & -1.10\%                  \\\hdashline
        OnPG            & 65.97$\pm$0.61\%   & -0.26\%   & 66.40$\pm$1.48\% & 0.10\%   \\
        OffPG           & \bf68.42$\pm$0.61\%    & \bf3.19\%    & \bf69.70$\pm$1.46\%    & \bf3.40\%\\
    \end{tabular}
    }
\end{table}

\begin{table}[t]
    \caption{Human evaluation side-by-side comparisons against the baseline. Positive values denote superior performance compared to the baseline.
    Note that some score increases for \Baseline+(t)~and \OnPG~are within error range.}
    \label{tab:sxs}
    \centering
    \scalebox{0.8}{
    \begin{tabular}{l|ccc}
                        & Informativeness & Correctness & Fluency                 \\ \hline\hline
        Baseline       & ---    & ---    & ---                    \\\hdashline
        Baseline+ (0.5)  & 1.78$\pm$0.85\%  & 0.18$\pm$0.49\%  & 0.10$\pm$0.28\%                  \\
        Baseline+ (0.7)  & 0.70$\pm$0.58\%	& 0.68$\pm$0.33\%	& 0.23$\pm$0.15\%                  \\\hdashline
        OnPG            & -0.33$\pm$0.90\%   & -0.35$\pm$0.62\%   & 0.08$\pm$0.20\% \\
        OffPG           & \bf7.45$\pm$1.06\%    & \bf5.90$\pm$0.80\%    & \bf1.69$\pm$0.31\% \\
    \end{tabular}
    }
\end{table}

\subsection{Evaluations}
\label{sec:exp-eval}
We run two sets of human evaluation studies to evaluate the performance of our models and baselines, using the T2 dataset (1K images). 
For every evaluation, we generate captions using beam search (beam size of 5). 

\subsubsection{Single-caption evaluation}
In the first type of evaluation, 6 distinct raters are asked to judge each image caption as good or bad. 
They are shown the image and caption with the ``Goodness'' question prompt shown in Table~\ref{tab:questions}. 
The bad or good rating is translated to 0 or 1, respectively. 
We measure ``average'' goodness score as the average of all the ratings over the test set.
We also report a ``voting''\footnote{The ``voting'' score is the metric reported on the Conceptual Captions Challenge leaderboard.} score which is the average of the binarized score for each caption based on majority voting. 
Note that both the ``average'' and ``voting'' scores are in the range $[0, 1]$, where higher values denote better model performance.

\subsubsection{Side-by-side caption evaluation}
In the other type of evaluation, we measure the relative improvement of a model against the \Baseline~model;
Three professional raters are shown the input image and two captions (anonymized and randomly shuffled with respect to their left/right position) side-by-side.
One of the captions is from a candidate model and the other always from \Baseline.
We ask for relative judgments on three dimensions -- Informativeness, Correctness and Fluency, using their corresponding questions shown in Table~\ref{tab:questions}.
Each of these dimensions allows a 5-way choice, shown below together with their corresponding scores:
\begin{center}
    \scalebox{0.8}{
    \begin{tabular}{lc}
        The left caption is much better        & $-1.0$ \\
        The left caption is slightly better    & $-0.5$ \\
        The two captions seem equal                 & ~~~$0.0$ \\
        The right caption is slightly better & $+0.5$ \\
        The right caption is much better     & $+1.0$ \\
    \end{tabular}
    }
\end{center}
Each model is evaluated by the average rating scores from 3 distinct raters.
As a result, we obtain 3 values for each model in the range $[-1, 1]$, where a negative score means a performance degradation in the given dimension with respect to \Baseline.
For every human evaluation, we report confidence intervals based on bootstrap resampling~\cite{koehn2004statistical}.

\begin{figure}[t]
    \centering
    \includegraphics[width=0.95\linewidth]{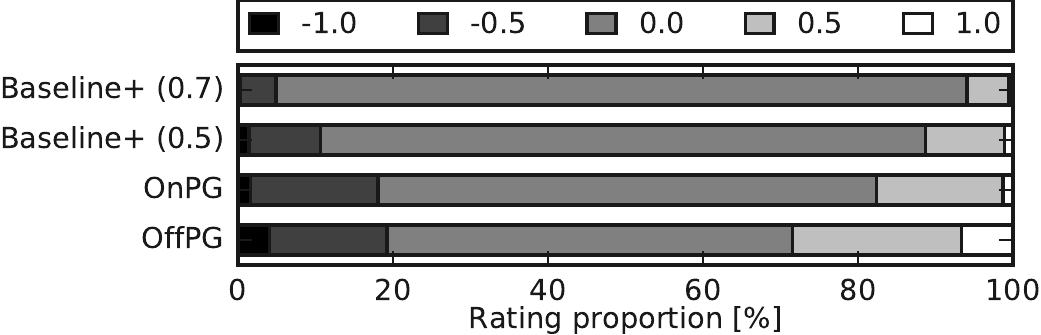}
    \caption{
    Rating distribution for the correctness question.
    Tendencies are similar for the other side-by-side questions.
    }
    \label{fig:rating_dist}
\end{figure}

\begin{figure}[t]
    \centering
    \includegraphics[width=0.9\linewidth]{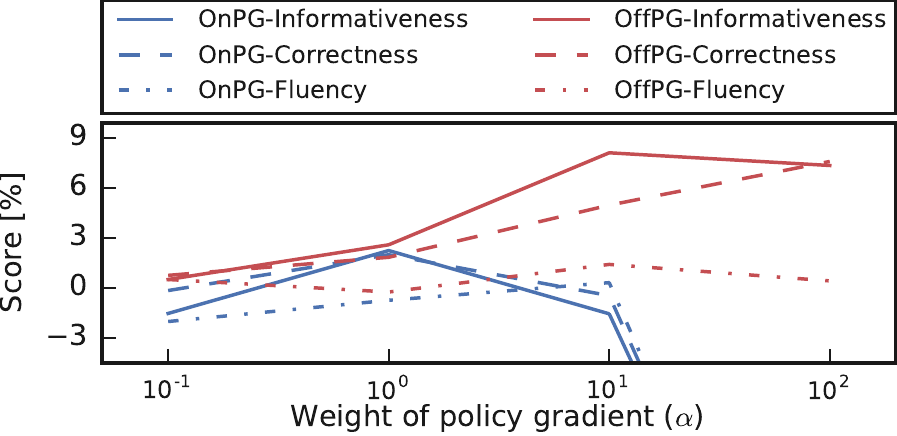}
    \caption{
    Results of \OnPG~and \OffPG~in side-by-side human comparisons while varying weight of policy gradient $\alpha$.
    Models are tested on 200 samples from T2 dataset.
    }
    \label{fig:varying_alpha}
\end{figure}

\begin{figure*}[!ht]
    \centering
    \includegraphics[width=\linewidth]{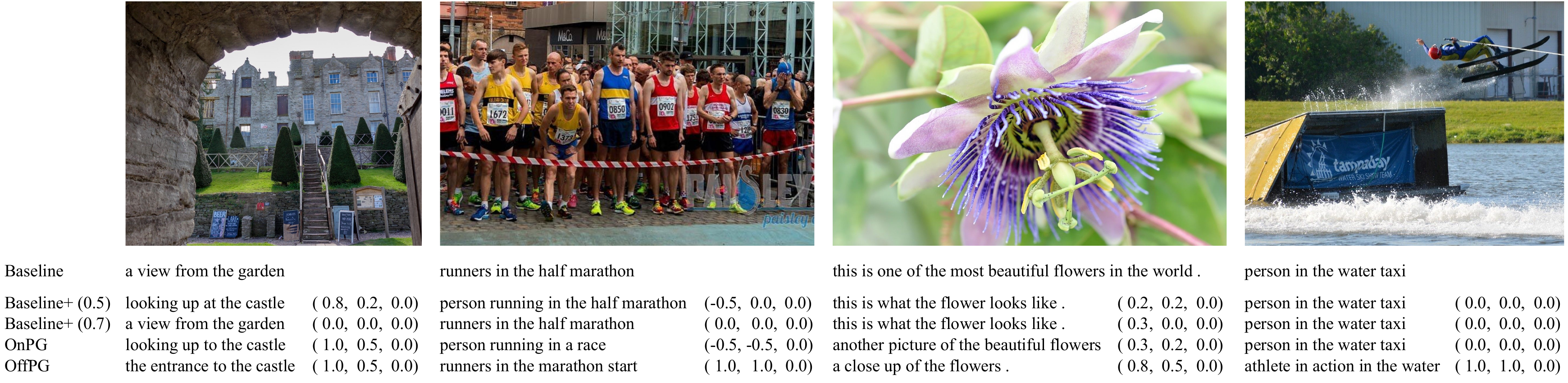}
    \caption{
    Qualitative examples of generated captions. 
    Numbers represent informativeness, correctness and fluency scores (rated by comparing against those generated by \Baseline).
    }
    \label{fig:qualitative}
\end{figure*}

\subsection{Results}

\subsubsection{Single-caption evaluation}
Table~\ref{tab:goodness} shows the goodness scores from the single-caption evaluation.
Both ``average'' and ``voting'' metrics clearly indicate that \OffPG~significantly improves over \Baseline, while the other methods achieve only marginal gains, all of which are within the error range.
\BaselinePlus{t} models use only 1.5\% and 2.2\% additional data, at $t=0.7$ and $t=0.5$, respectively, with insignificant impact.
Moreover, these methods only maximize the likelihood of the additional captions, which are already generated with high likelihood by previous models trained on the same dataset, which results in self-reinforcement.
In contrast, the policy gradient methods are allowed to utilize the negative feedback to directly penalize incorrect captions.
However, \OnPG~fails to improve the quality, most likely because it relies on a noisy caption ratings estimator that fails to generalize well over the large space of possible captions.

\subsubsection{Side-by-side evaluations}
The results from the side-by-side evaluations are are shown in Table~\ref{tab:sxs}.
The \OffPG~method achieves significant improvements on all three different dimensions.
This is an important result, considering that we trained the model using a caption ratings dataset that contains single-scalar  scores for generic 'goodness' (as opposed to the well-defined dimensions along which the \OffPG~method scores have improved).
These results demonstrate that the single-caption 'goodness' ratings encapsulate a signal for all these dimensions into its scalar value.
Note that we observe the same tendency consistently under a variety of hyperparameter settings in our internal experiments.

Figure~\ref{fig:rating_dist} highlights the way in which the \OffPG~method achieves its superiority over the \Baseline~model, compared to the other alternative models (using the 'Corectness' scores).
For instance, over 75\% of the captions for both \BaselinePlus{t}~models receive a 0.0 score (equal quality), and more than half of them are exactly identical to their corresponding \Baseline~captions.
In contrast, \OffPG~makes a strong impact by explicitly penalizing the captions with negative feedback:
less than 16\% captions are identical to the corresponding \Baseline~captions.
Moreover, we observe a large portion of captions with scores of 1.0 in favor of \OffPG, indicating that many captions are significantly enhanced.
We observe similar trends in all the three metrics. 

\subsubsection{On-policy vs. off-policy performance}
We compare the \OnPG~and \OffPG~methods in more depth, by performing ablation experiments for the $\alpha$ hyper-parameter (the weight for the policy gradient).
Figure~\ref{fig:varying_alpha} shows the results of these ablation experiments, for which we performed side-by-side comparisons over a 200-image subset from the T2 dataset.
The results indicate that a very small $\alpha$ limits the impact of the additional signal for both models, since the regularization effect from the original loss term becomes too strong.
By allowing updates using policy gradient with a larger $\alpha$ value, \OffPG~improves the performances along all three dimensions, whereas the performance of \OnPG~starts degrading at higher $\alpha$ values.
At $\alpha=100$, \OnPG~drastically suffers from mode collapse and ends up generating a single caption for every image.
This mode collapse is a result of poor generalization of the rating estimator: the collapsed captions are structurally ill-formed (\eg, an empty string, or a string with simply a period `.'), but they receive high rating estimates ($>0.9$) from the estimator.
Although we can (and did) introduce some heuristics to avoid some of these failure cases in the estimator, we observe that \OnPG~training would continue to suffer from the estimator failing to generalize well over the vast space of possible captions.
This observation is similar to the mode collapsing phenomenon seen when training generative adversarial networks (GANs), but even more severe as the estimator in \OnPG~is fixed (unlike the discriminators in GANs which are trained simultaneously).


Another drawback of \OnPG{} is that it increases the computational complexity significantly during training.
In terms of the memory usage, the rating estimator introduces 65\% additional parameters, and uses more than double the memory for gradient computation compared to the other models.
Also, the sequential caption sampling in \OnPG{} slows down the training procedure, by breaking the parallelism in the Transformer computations, in addition to the time complexity incurred by the rating estimator.
Empirically, \OnPG{} is over 10 times slower than the others in processing the same number of examples in training.
In contrast, the time and space complexities of \OffPG{} remain the same as \Baseline{} and \BaselinePlus{t}, since the only difference is the use of scalar weights ($r(c|I)$ and $\eta$) to gradients of each caption likelihood ($\bigtriangledown_\theta \ln p_\theta(c|I)$), as shown in Figure~\ref{fig:methods_relation}.

\subsubsection{Qualitative results}
Figure~\ref{fig:qualitative} presents some qualitative example outputs for our models, showcasing the effectiveness of the \OffPG~method.
We observe that the \OffPG~model is often successful at correcting arbitrary qualifiers present in the baseline outputs (\eg, \textit{`half marathon'} and \textit{`most beautiful'} in the second and third examples, respectively).

\section{Conclusion}
\label{sec:conclusion}
In this paper, we describe how to train an improved captioning model by using a caption ratings dataset, which is often a natural by-product in the development process of image captioning models.
We show that an off-policy RL technique with an alternative sampling distribution successfully deals with the sparsity of information about the rating function, while an on-policy method has difficulties in obtaining an improved model, due to generalization issues of the ratings estimator.
While this conclusion may not be definitive, it is definitely an important result, and it also opens up additional lines of inquiry along the relative merits of these RL techniques. 

\bibliographystyle{aaai}
\bibliography{refs}
\end{document}